\documentclass[conference]{IEEEtran}
\usepackage{stmaryrd}
\usepackage{color}

\usepackage[utf8]{inputenc}
\usepackage[T1]{fontenc}
\usepackage{multicol}
\usepackage{multirow}
\usepackage[pdftex]{graphicx}
\usepackage{multirow}
\usepackage{subfig}
\usepackage{amssymb}
\usepackage {upgreek}

\usepackage{amsmath}
\usepackage{dsfont}
\usepackage{multirow}
\usepackage{stmaryrd}

\usepackage{amssymb,amsmath}

\usepackage{booktabs}

\ifCLASSINFOpdf
\else
\fi
\begin{document}
%
\title{A mixture of experts model for\\predicting persistent weather patterns}



%
\author{\IEEEauthorblockN{M. P\'erez-Ortiz\IEEEauthorrefmark{1},
P.A. Guti\'errez\IEEEauthorrefmark{2},
P. Tino\IEEEauthorrefmark{3}, 
C. Casanova-Mateo\IEEEauthorrefmark{4} and
S. Salcedo-Sanz\IEEEauthorrefmark{5}}
\IEEEauthorblockA{\IEEEauthorrefmark{1}Computer Laboratory, University of Cambridge (UK),
email: mp867@cam.ac.uk}
\IEEEauthorblockA{\IEEEauthorrefmark{2}Dpt. of Computer Science and Numerical Analysis, University of C\'ordoba (Spain)\\}
\IEEEauthorblockA{\IEEEauthorrefmark{3}School of Computer Science, University of Birmingham (UK)\\}
\IEEEauthorblockA{\IEEEauthorrefmark{4}Department of Civil Engineering: Construction, Infrastructure and Transport, Universidad Politécnica de Madrid, Madrid (Spain)}
\IEEEauthorblockA{\IEEEauthorrefmark{5}Department of Signal Processing and Communications, University of Alcalá (Spain)}}


\maketitle

\begin{abstract}
Weather and atmospheric patterns are often persistent. The simplest weather forecasting method is the so-called persistence model, which assumes that the future state of a system will be similar (or equal) to the present state. Machine learning (ML) models are widely used in different weather forecasting applications, but they need to be compared to the persistence model to analyse whether they provide a competitive solution to the problem at hand. In this paper, we devise a new model for predicting low-visibility in airports using the concepts of mixture of experts. Visibility level is coded as two different ordered categorical variables: cloud height and runway visual height. The underlying system in this application is stagnant approximately in 90\% of the cases, and standard ML models fail to improve on the performance of the persistence model. Because of this, instead of trying to simply beat the persistence model using ML, we use this persistence as a baseline and learn an ordinal neural network model that refines its results by focusing on learning weather fluctuations. The results show that the proposal outperforms persistence and other ordinal autoregressive models, especially for longer time horizon predictions and for the runway visual height variable.
\end{abstract}



%
\IEEEpeerreviewmaketitle

\section{Introduction}


The persistence model, commonly used for weather forecasting, assumes that the conditions at the time of the forecast will remain unchanged, i.e. if it is rainy today, the persistence model will predict that it will be rainy tomorrow as well.
This method works well when weather patterns change slowly or weather is in a steady state, such as during the summer season in the tropics. This not only applies to short-term forecasting, but also for long range weather conditions (e.g. monthly predictions). 
In the last years, rapid Artic warming and uncommonly stationary waves of the jet stream have been seen to favour these persistent weather patterns \cite{1748-9326-10-1-014005}.



%

In some cases, persistence models are ``hard to beat'' by more sophisticated weather forecasting methods because of the stagnant dynamic of the system. This can happen even when information about the previous states is included in the prediction model.
As an alternative to this problem, this paper proposes the use of a strategy referred to in the literature as mixture of experts (ME), based on the divide and conquer principle. ME approaches assign different regions of the problem space to different experts, which are then supervised and managed by a gating function. Usually, the experts and the gating function are learnt together in an optimisation framework. The first expert that we consider is the persistence model, which already successfully predicts the next state of the system for most cases. The second expert is a machine learning model that predicts the output of the system when the persistence model is not accurate. Our model does not attempt to simply beat the persistence model, but rather assumes this persistent behaviour as a baseline and complement its performance when more drastic changes occur.

The problem considered in this paper is that of predicting low-visibility events at airports. Air transportation is probably the most affected sector by foggy and misty periods, since these events can dramatically restrict airport use and cause flight delays, diversions and cancellations \cite{Rebollo14}, or accidents in the worst cases \cite{Ahmed14}. 
The meteorological services that support air navigation systems prepare and disseminate terminal aerodrome forecasts to support the aeronautical community when dealing with airport low-visibility conditions.
These forecasts are used for pre and intra-flight planning and can help air traffic managers to activate procedures to ensure safe air operations during these events. 
 However, forecasting low-visibility conditions is not an easy task, mainly because fog formation is very sensitive to small-scale variations of atmospheric variables. For this reason, aeronautical meteorological forecasters need normally to integrate different sources of information to provide a robust prediction of low-visibility events. Recently, machine learning methods are gaining popularity for this task and they are being used to help forecasters improve the prediction of reduced visibility events at airports facilities \cite{Colabone15,Dutta15,en9080607}.

We evaluate the performance of our ME machine-learning model to forecast the runway visual range and cloud height at
the airport, both of which are crucial variables to determine low-visibility conditions. Contrary to previous approaches, the problem is tackled as an ordinal classification problem by discretising the time series in different categories. Given that four categories or ranges are enough for obtaining practical information, the main advantage of using this discretisation is the corresponding simplification of the prediction problem. The order of the categories implies the use of ordinal classifiers \cite{Gutierrez2016}, which are specifically designed for minimising the deviation of the predicted categories from the actual ones.

The methodology is tested with data collected at the Valladolid airport (Spain), which shows a persistence of approximately 90\% for hourly prediction. We evaluate the performance of our models at different time spans and with different window sizes. The prediction is performed using 7 atmospheric variables. Our results show that persistence model can be successfully complemented by machine learning, leading to a superior performance, specifically for long-term prediction, which is indeed necessary for successful airport managing.
 

The rest of the paper is structured as follows: Section \ref{sec:method} introduces the proposed mixture of experts, Section \ref{sec:exp} describes the datasets used, the experiments performed and the results obtained, and finally, Section \ref{sec:conclusions} outlines some conclusions and future work.

\section{Methodology} \label{sec:method}

This section presents the mixture of experts model proposed.

\subsection{Previous notions}


Our dataset $D$ is composed of a set of weather-related exogenous variables $\mathbf{x}$ and an output label $y$ which contains information about airport visibility, so that $D = \{(\mathbf{x}_1,y_1), \ldots, (\mathbf{x}_N,y_N)\}$. 
We study the prediction of runway visual range (RVR) and cloud height (CH), both of which are further described in Section \ref{sec:data}. These variables are discretised in classes, given that the application at hand might only requires a coarse-grain prediction and this greatly simplifies the prediction task. A common way of discretising a variable is:
\begin{equation}\label{eq:discretise}
y_t=\begin{cases}
\mathcal{C}_1, &\text{if } -\infty < r_t < R_1,\\
\mathcal{C}_2, &\text{if } R_1 \le r_t < R_2,\\
...\\
\mathcal{C}_Q, &\text{if } R_{Q-1} \le r_t < -\infty,
\end{cases}
\end{equation}
where $r_t$ is the real value being observed, and $R_1,\ldots,R_{Q-1}$ are a set of thresholds defined by the experts. The classes then
are known to follow a specific order of the form $\mathcal{C}_1 \prec \mathcal{C}_2 \prec \ldots \prec \mathcal{C}_q$, which can be tackled by a machine learning paradigm know as ordinal classification. This order derives a corresponding misclassification cost, since one aims to minimise the misclassification between classes further apart in the scale. 
 
\subsection{Proposed mixture of experts}

Our aim is to learn a model that complements the persistence in the cases when weather patterns are not stationary. Although we propose a probabilistic model, with a different probability equation for each class, we will start defining it as a regression model to ease its understanding.
The mixture of experts that we look for takes the following the form: 
\begin{equation}\label{eq1}
y_{t+k} = \alpha(\mathbf{z}_t)\cdot f_1(\mathbf{z}_t) + (1-\alpha(\mathbf{z}_t))\cdot f_2(\mathbf{z}_t), \;\; k \geq 1
\end{equation}
where the function $\alpha$ is a probabilistic gating function that decides whether we want to use $f_1$ or $f_2$ (i.e. the experts) to predict the output at $y_{t+k}$, and $k$ is the prediction horizon. The prediction is made based on $\mathbf{z}_t$, which we define as $\{\mathbf{x}_{t-\Delta},y_{t-\Delta},\ldots,\mathbf{x}_t,y_t\}$, where $\Delta$ is the window size, i.e. the number of instants we use for the prediction. 

In our case, since we know that the underlying dynamic of our system is mostly stationary, we define $f_1$ as the persistence model, i.e. $\hat{y}_{t+k} = f_1(\mathbf{z}_t) = y_t$. $f_2$ is then defined as an expert on the patterns that $f_1$ fails to predict. Eq. \ref{eq1} can be thus rewritten as: 
\begin{equation}
y_{t+k} = \alpha(\mathbf{z}_t)\cdot y_t + (1-\alpha(\mathbf{z}_t))\cdot f_2(\mathbf{z}_t),\label{eq:regressionModel}
\end{equation}
where $y_t$ is known and $\alpha$ and $f_2$ needs to be optimised. We define $f_2$ as an autoregressive neural network and $\alpha$ as an autoregressive logistic regression function. Both functions can be optimised together through gradient descent.

%
%

Given that we are dealing with a classification problem, we need to adapt Eq. \ref{eq:regressionModel} and separately estimate the probability that a pattern $y_{t+k}$ belongs to $\mathcal{C}_q$:
\begin{eqnarray}\label{eq:Prob}
&P(y_{t+k}=\mathcal{C}_q|\mathbf{z}_t,{\boldsymbol\upnu},{\boldsymbol\upkappa}) = \alpha(\mathbf{z}_t,{\boldsymbol\upnu})\cdot [[y_t = \mathcal{C}_q]] + \ldots \\ &+(1-\alpha(\mathbf{z}_t,{\boldsymbol\upnu}))\cdot P_{\texttt{net}}(y_{t+k} = \mathcal{C}_q| \mathbf{z}, {\boldsymbol\upkappa}), \nonumber
\end{eqnarray}
where $[[\cdot]]$ is a boolean test returning $1$ if the condition is true ($0$ otherwise) and represents the persistence for each individual probability, $P_{\texttt{net}}(y_{t+k} = \mathcal{C}_q| \mathbf{z}_t, {\boldsymbol\upkappa})$ is a probabilistic autoregressive neural network (which will be further detailed in Section \ref{sec:nnpom}) with parameters ${\boldsymbol\upkappa}$, and $\alpha(\mathbf{z}_t,{\boldsymbol\upnu})$ is a probabilistic autoregressive logistic regression model with parameters ${\boldsymbol\upnu}$, i.e.:
\begin{equation*}
\alpha(\mathbf{z}_t,{\boldsymbol\upnu}) = \frac{1}{1+\exp(-{\boldsymbol\upnu}^T \cdot (1,\mathbf{z}_t))}=\sigma({\boldsymbol\upnu}^T \cdot (1,\mathbf{z}_t)),
\end{equation*}
where $\sigma(x)$ is the sigmoid function. The class predicted will be given the maximum a posteriori probability:
\begin{eqnarray*}
\hat{y}_{t+k} = \arg\max_{\mathcal{C}_q} P(y_{t+k}=\mathcal{C}_q|\mathbf{z}_t,{\boldsymbol\upnu},{\boldsymbol\upkappa}).
\end{eqnarray*}

\subsection{Learning the free parameters}

We propose two schemes for optimising the parameters ${\boldsymbol\upkappa}$ and ${\boldsymbol\upnu}$, which are detailed below.

\subsubsection{Independent training}\label{sec:indepModel}

The most simple strategy for optimising these parameters is to train them independently following these steps: 
\begin{enumerate} 
	\item Run $\hat{y}_{t+k} = y_t$ (i.e. compute $f_1$) through $D$.
	\item Identify problematic cases $\mathbf{Z} = \{\mathbf{z}_{n_1}, \ldots, \mathbf{z}_{n_\mathrm{p}}\}$, i.e. patterns for which the persistence model does not predict the result accurately such that $y_{t+k} \neq y_t$.
	\item Define $C = \{(\mathbf{z}_1,c_1), \ldots, (\mathbf{z}_N,c_N)\}$, where $c_i$ is defined as $0$ if the pattern is problematic, and $1$ otherwise.
	\item Learn $\alpha(\mathbf{z}_t,{\boldsymbol\upnu})$ using $C$, which is a binary problem. We consider a standard logistic regression learner.
	\item Train $f_2$ only on problematic cases, $P= \{(\mathbf{z}_{n_1},y_{n_1+k}), \ldots, (\mathbf{z}_{n_1},y_{n_\mathrm{p}+k})\}$. The error function and the neural network model used are similar to the ones explained in the next subsection.
\end{enumerate}

This strategy focuses the neural network component only on problematic cases and does not consider potential interactions between both the logistic regression model and the network.

\subsubsection{Simultaneous training}\label{sec:simulModel}

To obtain the best potential of this mixture of experts, it is desirable to optimise both models $\alpha$ and $f_2$ together.
A convenient strategy is thus to apply gradient descent over the whole parameter vector $\mathbf{s}=\{{\boldsymbol\upnu},{\boldsymbol\upkappa}\}$
with the training set $D$.

The cross-entropy function can be minimised for this purpose:
\begin{eqnarray*}
L_\mathrm{O}(\mathbf{s},D) = -\frac{1}{N} \sum _{t=1}^{N}\sum _{q=1}^{Q}[[y_{t+k} = \mathcal{C}_q]] \log p_{tq},
\end{eqnarray*}
where $p_{tq}=P(y_{t+k}=\mathcal{C}_q|\mathbf{z}_t,{\boldsymbol\upnu},{\boldsymbol\upkappa})$ is the estimation of probability given by Eq. \ref{eq:Prob}. The gradient descent technique used is the $iRprop^{+}$ algorithm, which usually provides robust performance \cite{Igel2003}.

Given that the datasets are moderately imbalanced (see Section \ref{sec:data}), we also include different costs for the classes of the problem, according to their a priori probability:
\begin{eqnarray}
L_\mathrm{W}(\mathbf{s},D) = -\frac{1}{N} \sum _{t=1}^{N} \sum _{q=1}^{Q} o_q[[y_{t+k} = \mathcal{C}_q]] \log p_{tq}, \nonumber
\end{eqnarray}
where $o_q=1-\frac{N_q}{N}$, and $N_q$ is the number of patterns of class $\mathcal{C}_q$. We also include $L2$ regularisation in the error function to avoid overfitting, so that the final cost is:
\begin{eqnarray}\label{eq:entropia3}
L(\mathbf{s},D) = L_\mathrm{W}(\mathbf{s},D)+ \lambda \cdot \sum_{i=1}^S s_i^2,
\end{eqnarray}
where $S$ is the total number of parameters and $\lambda$ is the regularisation parameter.

Now we detail the derivatives of the error function with respect to the network parameters. The gradient vector is given by:
\begin{equation*}
\nabla L(\mathbf{s},D) = \left(\frac{\partial L(\mathbf{s},D)}{\partial s_1},\frac{\partial L(\mathbf{s},D)}{\partial s_2},\ldots,
\frac{\partial L(\mathbf{s},D)}{\partial s_S} \right).
\end{equation*}

Considering Eq. \ref{eq:entropia3}, each of the components can be defined as:
\begin{eqnarray*}
\frac{\partial L}{\partial s_i}= -\frac{1}{N} \sum _{t=1}^{N}\sum _{q=1}^{Q} \frac{o_q[[y_{t+k} = \mathcal{C}_q]]}{p_{tq}}\cdot\frac{\partial {p_{tq}}}{\partial s_i} + 2s_i,\nonumber
\end{eqnarray*}
where $i=1,\ldots,S$. For the logistic regression parameters, these derivatives are given by:
\begin{eqnarray*}
\frac{\partial p_{tq}}{\partial \nu_i} & = &  \sigma'(h)[[y_{t} = \mathcal{C}_q]]-\sigma'(h)p_{\mathrm{net}nq}, \nonumber
\end{eqnarray*}
where $h={\boldsymbol\upnu}^T \cdot (1,\mathbf{z})$, $\sigma'(x)=\sigma(x)(1-\sigma(x))$, and $p_{\mathrm{net}nq}=P_{\texttt{net}}(y_{t+k} = \mathcal{C}_q| \mathbf{z}_t, {\boldsymbol\upkappa})$. The specific model for $p_{\mathrm{net}nq}$ and the corresponding derivatives will be detailed in next subsection.

\subsection{Proportional Odds Model Neural Network (NNPOM)}\label{sec:nnpom}
This section defines the neural network component in our model ($f_2$). 
As stated before, labels are ordered, which encourages the use of an ordinal classification model \cite{gutierrez2016current}. In this paper, we adopt a probabilistic framework and consider the Proportional Odds Model Neural Network (NNPOM) presented in previous research \cite{Gutierrez2014,Mathieson1996}. NNPOM is a threshold model, i.e. it approaches ordinal classification by trying to estimate the latent variable originating the different ordinal categories and learning a set of thresholds discretising this variable. Threshold models have been seen to perform well when categories come from a discretised variable \cite{Gutierrez2016}.

NNPOM extends the Proportion Odds Model (POM) \cite{McCullagh1980}, which, at the same time, is an extension of binary logistic regression. NNPOM  predicts cumulative probabilities $P(y_{t+k}\preceq\mathcal{C}_j|\mathbf{z}_t)$, which can be used to obtain direct probability estimations as:
\begin{align}\label{eq:Probabilities}
P(y_{t+k}\preceq\mathcal{C}_q|\mathbf{z}_t) & =\sum_{j=1}^q P(y_{t+k}=\mathcal{C}_j|\mathbf{z}_t),\nonumber\\
P(y_{t+k}=\mathcal{C}_q|\mathbf{z}_t) & = P(y_{t+k}\preceq\mathcal{C}_q|\mathbf{z}_t)-\\
&- P(y_{t+k}\preceq\mathcal{C}_{q-1}|\mathbf{z}_t),\nonumber
\end{align}
{with $q=2, \ldots, Q$, and considering by definition that $P(y_{t+k}=\mathcal{C}_1|\mathbf{z}_t)=P(y_{t+k}\preceq\mathcal{C}_1|\mathbf{z}_t)$ and $P(y_{t+k}\preceq\mathcal{C}_Q|\mathbf{z}_t)=1$}.
NNPOM assumes a logistic function for the distribution of the random error component of the latent variable, giving rise to:
\begin{eqnarray}\label{eq:Probabilities2}
P(y_{t+k}\preceq\mathcal{C}_q|\mathbf{z}_t)=\sigma(f(\mathbf{z}_t,{\boldsymbol \uptheta})-b_q),
\end{eqnarray}
where $q=2, \ldots, Q-1$, $b_q$ is the threshold for class $\mathcal{C}_q$, and $f(\mathbf{z}_t,{\boldsymbol \uptheta})$ is the projection of the model for pattern $\mathbf{z}_t$. The following constraints must be satisfied: $b_1\le\ldots\le b_{Q-1}$, in order to ensure the monotonicity of the cumulative probabilities. However, we can apply unconstrained optimisers by defining the thresholds as:
\begin{equation*}
b_{q} = b_1 + \sum_{j=2}^{q} a_j^2
\end{equation*}
with padding variables $a_j$, which are squared to make them positive, and $b_1,a_j \in \mathbb{R}$. 

\begin{figure}
\centering
\includegraphics[keepaspectratio,width=.4\textwidth]{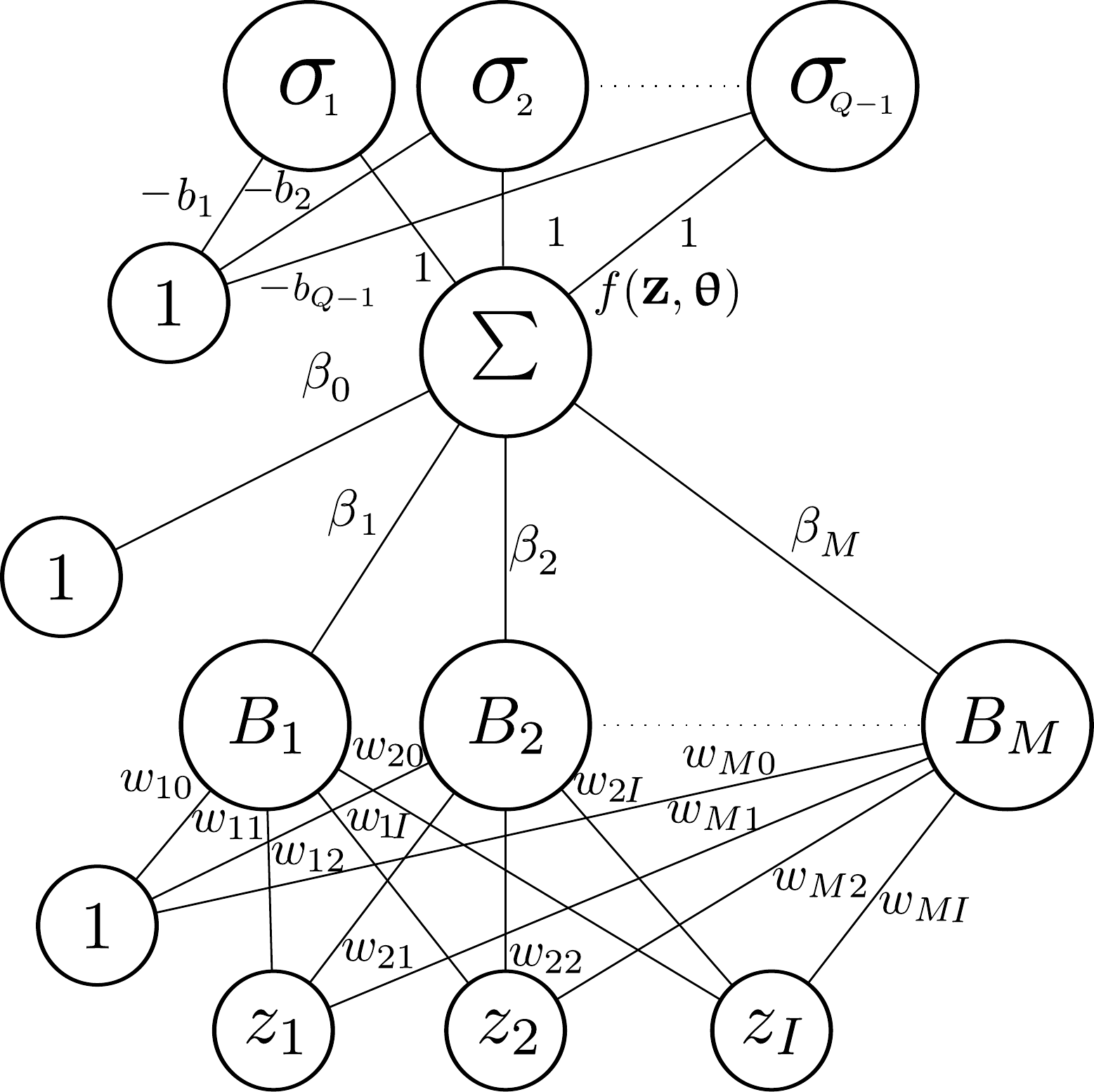}
\caption{Structure of the NNPOM model, including one output node with different biases, $M$ hidden nodes and $k$ input nodes}\label{fig:NNPOM}
\end{figure}

Considering Eq. \ref{eq:Probabilities} and \ref{eq:Probabilities2}, the probabilities estimated by POM and NNPOM are defined in the following way:
\begin{align}
P(y_{t+k}=\mathcal{C}_1|\mathbf{z}_t)=& \sigma(f(\mathbf{z}_t,{\boldsymbol \uptheta})-b_1),\nonumber\\
P(y_{t+k}=\mathcal{C}_q|\mathbf{z}_t)= &
\sigma(f(\mathbf{z}_t,{\boldsymbol \uptheta})-b_q)-\sigma(f(\mathbf{z}_t,{\boldsymbol \uptheta})-b_{q-1}),\nonumber\\
& q\in\{2,\ldots,Q-1\},\nonumber\\
P(y_{t+k}=\mathcal{C}_Q|\mathbf{z}_t)=& 1-\sigma(f(\mathbf{z}_t,{\boldsymbol \uptheta})-b_{Q-1}).\nonumber
\end{align}
The main difference between POM and NNPOM is found in $f(\mathbf{z}_t,{\boldsymbol \uptheta})$: POM estimates the latent variable as a linear model of the inputs, $f(\mathbf{z}_t,{\boldsymbol \uptheta})={\boldsymbol \uptheta}^T\cdot(1,\mathbf{z}_t)$, while a linear combination of nonlinear basis functions (hidden neurons) is assumed for NNPOM:
\begin{equation}
f(\mathbf{z}_t,{\boldsymbol \uptheta})=\sum_{j=1}^{M}\beta_{j}B_j(\mathbf{z}_t,\mathbf{w}_{j}),\nonumber
\end{equation}
where $M$ is the number of hidden units, ${\boldsymbol \uptheta}=\{{\boldsymbol \upbeta},\mathbf{W}\}$, ${\boldsymbol \upbeta}=\{\beta_1,\ldots,\beta_M\}$, $\mathbf{W}=\{\mathbf{w}_1,\ldots,\mathbf{w}_M\}$, and $B_j(\mathbf{z}_t,\mathbf{w}_{j})$ can be any type of basis functions, in our case, sigmoidal units, $B_j(\mathbf{z}_t,\mathbf{w}_{j})=\sigma(\mathbf{w}_j^T\cdot\left(1,\mathbf{z}_t\right))$, where $\mathbf{w}_j=\{w_{j0},w_{j1},\ldots,w_{jI}\}$, $I$ is the number of inputs and $w_0$ is the bias. The final parameter vector of NNPOM 
is defined as ${\boldsymbol\upkappa}=\left\{\boldsymbol{\upbeta},\mathbf{W},b_1,a_2,\ldots,a_{Q-1}\right\}$. The structure of the NNPOM model is detailed in \figurename{ \ref{fig:NNPOM}}.

To ease the notation, we define $p_{tq}=P_\mathrm{net}(y_{t+k}=\mathcal{C}_q|\mathbf{z}_t,{\boldsymbol\upkappa})$, $g_{tq}=g(\mathbf{x}_t,{\boldsymbol\upkappa})=\sigma(f_t-b_q)$ and $f_{t}=f(\mathbf{x}_t,{\boldsymbol \upbeta},\mathbf{W})$. According to Eq. \ref{eq:Probabilities}, the derivatives of the probability function can be specified by:
\begin{eqnarray}
\frac{\partial p_{t1}}{\partial s_i} & = &  \sigma'(g_{t1})\frac{\partial g_{t1}}{\partial s_i}, \nonumber\\
\frac{\partial p_{tq|1<q<Q}}{\partial s_i} & = &  \sigma'(g_{tq})\frac{\partial g_{tq}}{\partial s_i} - \sigma'(g_{t(q-1)})\frac{\partial g_{t(q-1)}}{\partial s_i},\nonumber \\
\frac{\partial p_{tQ}}{\partial s_i} & = &  -\sigma'(g_{t(Q-1)})\frac{\partial g_{t(Q-1)}}{\partial s_i},\nonumber
\end{eqnarray}
where $s_i\in{\boldsymbol\upkappa}$.

The derivatives for $g_{tq}$ (i.e. $s_i\in\{{\boldsymbol \upbeta},\mathbf{W}\}$) are the standard ones for multilayer perceptrons:
\begin{eqnarray}
	\frac{\partial g_{tq}}{\partial \beta_{j}}=\frac{\partial f_{t}}{\partial \beta_{j}} & = &  B_j(\mathbf{z}_t,\mathbf{w}_j), \nonumber \\
	\frac{\partial g_{tq}}{\partial w_{ji}}=\frac{\partial f_{t}}{\partial w_{ji}} & = & \beta_{j}  \sigma'(\mathbf{w}_j^T\cdot\left(1,\mathbf{z}_t\right)) z_{ti},\nonumber
\end{eqnarray}
where $i\in\{1,\ldots,I\}$, $j\in\{1,\ldots,M\}$, and $I$ is the number of input variables.

For the threshold and padding parameters, the derivatives are:
\begin{eqnarray}
\frac{\partial g_{tq}}{\partial b_1}=-1, \frac{\partial g_{tq}}{\partial a_j}=\begin{cases}
0, & \text{if }q<j,\\
-2a_j, & \text{otherwise}.
\end{cases} \nonumber
\end{eqnarray}
where $j\in\{2,\ldots,Q-1\}$.

\begin{table}[t!]
	\caption{Number of patterns and class distribution of the datasets for different time horizons ($k$) and window sizes ($\Delta$).}
	\centering
	\begin{tabular}{rcc}
		\toprule
		\multicolumn{3}{c}{Dataset: CH}\\
		\midrule
		& \# Patterns & Distribution\\
		\midrule
		
		$\Delta=1$, $k=1$ & $5974$ & $[4178,1170,626]$ \\
		$\Delta=1$, $k=3$ & $5261$ & $[3749,1017,495]$ \\
		$\Delta=1$, $k=6$ & $4200$ & $[3156,747,297]$ \\
		$\Delta=3$, $k=1$ & $5258$ & $[3746,1017,495]$ \\
		$\Delta=3$, $k=3$ & $4549$ & $[3350,848,351]$ \\
		$\Delta=3$, $k=6$ & $3489$ & $[2751,520,218]$ \\
		$\Delta=5$, $k=1$ & $4546$ & $[3347,848,351]$ \\
		$\Delta=5$, $k=3$ & $3837$ & $[2957,628,252]$ \\
		$\Delta=5$, $k=6$ & $2779$ & $[2254,351,174]$ \\
		\midrule
		\multicolumn{3}{c}{Dataset: RVR}\\
		\midrule
		& \# Patterns & Distribution\\
		\midrule
		$\Delta=1$, $k=1$ & $8520$ & $[7199,903,312,106]$ \\
		$\Delta=1$, $k=3$ & $8518$ & $[7197,903,312,106]$ \\
		$\Delta=1$, $k=6$ & $8515$ & $[7194,903,312,106]$ \\
		$\Delta=3$, $k=1$ & $8518$ & $[7197,903,312,106]$ \\
		$\Delta=3$, $k=3$ & $8516$ & $[7195,903,312,106]$ \\
		$\Delta=3$, $k=6$ & $8513$ & $[7192,903,312,106]$ \\
		$\Delta=5$, $k=1$ & $8516$ & $[7195,903,312,106]$ \\
		$\Delta=5$, $k=3$ & $8514$ & $[7193,903,312,106]$ \\
		$\Delta=5$, $k=6$ & $8511$ & $[7190,903,312,106]$ \\
		\bottomrule
	\end{tabular}
	\label{tab:datasets}
\end{table}

\section{Experiments}\label{sec:exp}

This section presents the performance of the previously presented approaches for low-visibility events prediction and analyses the results obtained.

\begin{table*}[t!]
	\caption{Mean test results obtained by the different methods compared for CH and RVR and different time horizons (k) and window size ($\Delta$).}
	\centering
	\begin{tabular}{rrccccrrcccc}
		\toprule
		Dataset: CH & Method & $Acc$ & $AMAE$ & $MMAE$ & $GM$ & Dataset: RVR & Method & $Acc$ & $AMAE$ & $MMAE$ & $GM$ \\
		\midrule
		$\Delta=1$, $k=1$ & Persist & $\mathit{87.36}$ & $\mathbf{0.2087}$ & $\mathit{0.2866}$ & $\mathbf{81.54}$ & $\Delta=1$, $k=1$ & Persist & $89.80$ & $\mathbf{0.3776}$ & $\mathbf{0.5711}$ & $\mathbf{63.34}$\\
		& POM & $87.16$ & $0.2323$ & $0.3307$ & $78.79$ &  & POM & $\mathbf{90.76}$ & $\mathit{0.3808}$ & $\mathit{0.5746}$ & $\mathit{62.73}$\\
		& NNPOM & $86.52$ & $0.2519$ & $0.3817$ & $76.81$ &  & NNPOM & $90.55$ & $0.4489$ & $0.7927$ & $52.66$\\
		& ITME & $87.32$ & $\mathit{0.2091}$ & $\mathit{0.2866}$ & $\mathit{81.52}$ &  & ITME & $89.64$ & $0.4643$ & $0.9227$ & $55.35$\\
		& STME & $\mathbf{87.41}$ & $0.2340$ & $0.3475$ & $79.63$ &  & STME & $\mathit{90.66}$ & $0.4185$ & $0.6444$ & $58.76$\\
		& STMEIC & $85.93$ & $0.2130$ & $\mathbf{0.2832}$ & $81.31$ &  & STMEIC & $86.98$ & $0.3939$ & $0.6149$ & $61.77$\\
		\midrule
		$\Delta=1$, $k=3$ & Persist & $\mathbf{77.44}$ & $\mathbf{0.3759}$ & $\mathbf{0.5141}$ & $\mathbf{67.22}$ & $\Delta=1$, $k=3$ & Persist & $83.74$ & $\mathit{0.7239}$ & $\mathit{1.2927}$ & $\mathit{37.68}$\\
		& POM & $77.02$ & $0.5181$ & $0.9767$ & $35.63$ &  & POM & $\mathbf{85.88}$ & $0.9250$ & $1.8635$ & $7.42$\\
		& NNPOM & $74.98$ & $0.5719$ & $1.0173$ & $43.82$ &  & NNPOM & $85.29$ & $0.8594$ & $1.5876$ & $31.44$\\
		& ITME & $77.17$ & $\mathit{0.4168}$ & $\mathit{0.6290}$ & $\mathit{64.51}$ &  & ITME & $84.29$ & $0.9178$ & $1.8055$ & $21.04$\\
		& STME & $\mathit{77.32}$ & $0.5055$ & $0.8392$ & $55.45$ &  & STME & $\mathit{85.49}$ & $0.8583$ & $1.6031$ & $32.30$\\
		& STMEIC & $77.00$ & $0.4222$ & $0.7143$ & $61.43$ &  & STMEIC & $80.56$ & $\mathbf{0.6849}$ & $\mathbf{1.2764}$ & $\mathbf{38.19}$\\
		\midrule
		$\Delta=1$, $k=6$ & Persist & $67.95$ & $\mathbf{0.5000}$ & $\mathbf{0.6588}$ & $\mathbf{55.04}$ & $\Delta=1$, $k=6$ & Persist & $79.94$ & $0.9796$ & $1.8381$ & $23.13$\\
		& POM & $\mathit{75.36}$ & $0.7517$ & $1.4683$ & $0.00$ &  & POM & $\mathbf{84.46}$ & $1.1937$ & $2.4438$ & $0.00$\\
		& NNPOM & $\mathbf{75.45}$ & $0.6630$ & $1.2531$ & $34.94$ &  & NNPOM & $84.12$ & $\mathit{0.9296}$ & $1.7808$ & $27.93$\\
		& ITME & $72.31$ & $0.7285$ & $1.4201$ & $27.70$ &  & ITME & $84.09$ & $1.0603$ & $1.8855$ & $19.51$\\
		& STME & $73.86$ & $0.6302$ & $1.1380$ & $\mathit{42.75}$ &  & STME & $\mathit{84.18}$ & $0.9332$ & $\mathit{1.7448}$ & $\mathit{28.02}$\\
		& STMEIC & $70.48$ & $\mathit{0.5895}$ & $\mathit{1.0624}$ & $41.58$ &  & STMEIC & $78.55$ & $\mathbf{0.8049}$ & $\mathbf{1.5195}$ & $\mathbf{32.02}$\\
		\midrule
		$\Delta=3$, $k=1$ & Persist & $87.36$ & $0.2087$ & $0.2866$ & $81.54$ & $\Delta=3$, $k=1$ & Persist & $89.80$ & $\mathbf{0.3776}$ & $\mathbf{0.5711}$ & $\mathbf{63.33}$\\
		& POM & $87.02$ & $0.2370$ & $0.3395$ & $78.26$ &  & POM & $\mathbf{91.10}$ & $\mathit{0.3882}$ & $\mathit{0.5870}$ & $\mathit{62.66}$\\
		& NNPOM & $86.64$ & $0.2453$ & $0.3775$ & $76.45$ &  & NNPOM & $90.46$ & $0.4753$ & $0.8415$ & $51.10$\\
		& ITME & $\mathbf{87.53}$ & $\mathit{0.1959}$ & $\mathit{0.2728}$ & $\mathit{82.11}$ &  & ITME & $89.40$ & $0.5015$ & $1.0448$ & $50.03$\\
		& STME & $\mathit{87.48}$ & $0.2216$ & $0.3355$ & $79.72$ &  & STME & $\mathit{90.84}$ & $0.4345$ & $0.7053$ & $57.23$\\
		& STMEIC & $86.03$ & $\mathbf{0.1939}$ & $\mathbf{0.2469}$ & $\mathbf{82.28}$ &  & STMEIC & $86.43$ & $0.4159$ & $0.6864$ & $60.11$\\
		\midrule
		$\Delta=3$, $k=3$ & Persist & $77.43$ & $\mathit{0.3760}$ & $\mathbf{0.5141}$ & $\mathbf{67.22}$ & $\Delta=3$, $k=3$ & Persist & $83.74$ & $0.7240$ & $1.2927$ & $37.68$\\
		& POM & $77.41$ & $0.5041$ & $0.9228$ & $45.94$ &  & POM & $86.25$ & $0.8956$ & $1.8254$ & $7.74$\\
		& NNPOM & $\mathit{79.28}$ & $0.4653$ & $0.8050$ & $54.68$ &  & NNPOM & $\mathit{86.27}$ & $\mathit{0.6771}$ & $\mathit{1.2622}$ & $\mathbf{44.30}$\\
		& ITME & $76.71$ & $0.4810$ & $0.8944$ & $52.07$ &  & ITME & $84.75$ & $0.8585$ & $1.6475$ & $24.33$\\
		& STME & $\mathbf{79.81}$ & $0.4339$ & $0.7271$ & $60.38$ &  & STME & $\mathbf{86.60}$ & $0.6890$ & $1.3021$ & $41.70$\\
		& STMEIC & $77.16$ & $\mathbf{0.3556}$ & $\mathit{0.5476}$ & $\mathit{67.21}$ &  & STMEIC & $81.39$ & $\mathbf{0.6002}$ & $\mathbf{1.1624}$ & $\mathit{42.08}$\\
		\midrule
		$\Delta=3$, $k=6$ & Persist & $67.95$ & $\mathit{0.5000}$ & $\mathbf{0.6588}$ & $\mathbf{55.04}$ & $\Delta=3$, $k=6$ & Persist & $79.94$ & $0.9797$ & $1.8381$ & $23.13$\\
		& POM & $75.73$ & $0.7147$ & $1.3994$ & $12.45$ &  & POM & $84.47$ & $1.1285$ & $2.3229$ & $0.00$\\
		& NNPOM & $\mathbf{80.40}$ & $0.5284$ & $0.9666$ & $50.90$ &  & NNPOM & $84.31$ & $0.7569$ & $1.3973$ & $41.15$\\
		& ITME & $76.02$ & $0.6622$ & $1.2550$ & $32.95$ &  & ITME & $\mathit{84.52}$ & $0.9720$ & $1.7125$ & $27.03$\\
		& STME & $\mathit{79.51}$ & $0.5535$ & $0.9893$ & $47.64$ &  & STME & $\mathbf{85.14}$ & $\mathit{0.7347}$ & $\mathit{1.3259}$ & $\mathbf{43.41}$\\
		& STMEIC & $71.93$ & $\mathbf{0.4798}$ & $\mathit{0.7766}$ & $\mathit{52.74}$ &  & STMEIC & $78.29$ & $\mathbf{0.6632}$ & $\mathbf{1.2319}$ & $\mathit{43.16}$\\
		\midrule
		$\Delta=5$, $k=1$ & Persist & $\mathit{87.35}$ & $\mathbf{0.2088}$ & $\mathbf{0.2866}$ & $\mathbf{81.53}$ & $\Delta=5$, $k=1$ & Persist & $89.80$ & $\mathbf{0.3776}$ & $\mathbf{0.5711}$ & $\mathbf{63.33}$\\
		& POM & $86.73$ & $0.2417$ & $0.3448$ & $77.79$ &  & POM & $\mathbf{90.81}$ & $\mathit{0.4040}$ & $\mathit{0.6278}$ & $\mathit{60.64}$\\
		& NNPOM & $86.77$ & $0.2724$ & $0.4476$ & $72.89$ &  & NNPOM & $90.16$ & $0.4843$ & $0.8467$ & $51.15$\\
		& ITME & $\mathbf{87.92}$ & $\mathit{0.2139}$ & $\mathit{0.3076}$ & $\mathit{80.97}$ &  & ITME & $89.35$ & $0.5316$ & $1.1571$ & $46.47$\\
		& STME & $87.35$ & $0.2741$ & $0.4578$ & $75.13$ &  & STME & $\mathit{90.74}$ & $0.4398$ & $0.7318$ & $56.88$\\
		& STMEIC & $85.37$ & $0.2268$ & $0.3314$ & $78.97$ &  & STMEIC & $86.26$ & $0.4309$ & $0.7244$ & $58.19$\\
		\midrule
		$\Delta=5$, $k=3$ & Persist & $77.42$ & $\mathbf{0.3760}$ & $\mathbf{0.5141}$ & $\mathbf{67.21}$ & $\Delta=5$, $k=3$ & Persist & $83.73$ & $0.7240$ & $1.2927$ & $37.68$\\
		& POM & $77.71$ & $0.5023$ & $0.8952$ & $48.51$ &  & POM & $86.28$ & $0.8737$ & $1.7775$ & $8.04$\\
		& NNPOM & $\mathbf{81.80}$ & $0.4801$ & $0.8773$ & $54.26$ &  & NNPOM & $\mathit{87.10}$ & $\mathit{0.6217}$ & $\mathit{1.1599}$ & $\mathbf{50.13}$\\
		& ITME & $78.02$ & $0.5554$ & $1.1021$ & $39.50$ &  & ITME & $85.10$ & $0.8498$ & $1.6264$ & $26.04$\\
		& STME & $\mathit{81.36}$ & $0.4795$ & $0.8542$ & $53.76$ &  & STME & $\mathbf{87.26}$ & $0.6247$ & $\mathbf{1.1571}$ & $\mathit{48.72}$\\
		& STMEIC & $77.33$ & $\mathit{0.3786}$ & $\mathit{0.6406}$ & $\mathit{63.38}$ &  & STMEIC & $81.50$ & $\mathbf{0.6129}$ & $1.2036$ & $41.11$\\
		\midrule
		$\Delta=5$, $k=6$ & Persist & $67.95$ & $0.5000$ & $\mathbf{0.6588}$ & $\mathit{55.04}$ & $\Delta=5$, $k=6$ & Persist & $79.93$ & $0.9796$ & $1.8381$ & $23.13$\\
		& POM & $75.47$ & $0.7307$ & $1.4406$ & $12.41$ &  & POM & $84.90$ & $1.1021$ & $2.2790$ & $0.00$\\
		& NNPOM & $\mathit{82.78}$ & $0.5035$ & $0.9189$ & $52.64$ &  & NNPOM & $\mathit{85.10}$ & $0.7345$ & $1.3738$ & $\mathit{43.22}$\\
		& ITME & $77.46$ & $0.6819$ & $1.2438$ & $34.67$ &  & ITME & $84.40$ & $0.9835$ & $1.7932$ & $27.56$\\
		& STME & $\mathbf{83.53}$ & $\mathit{0.4818}$ & $0.8678$ & $54.80$ &  & STME & $\mathbf{85.58}$ & $\mathit{0.7204}$ & $\mathit{1.3403}$ & $\mathbf{45.40}$\\
		& STMEIC & $75.38$ & $\mathbf{0.4630}$ & $\mathit{0.7646}$ & $\mathbf{55.44}$ &  & STMEIC & $78.15$ & $\mathbf{0.6650}$ & $\mathbf{1.2552}$ & $42.72$\\
		\bottomrule
	\end{tabular}
	\label{tab:resultsGlobal}
\end{table*}

\subsection{Data description}\label{sec:data}

The datasets used consider the prediction of low-visibility events at Valladolid airport, Spain (41.70 N, 4.88 W). This airport is well-known for its foggy days. Due to its geographical and climatological characteristics, radiation fog is very frequent \cite{Roman-Gascon16}. A detailed Valladolid airport climatology can be found in \cite{AEMET12}.

In order to have in-situ information about the most basic parameters involved in radiation fog events at the airport, we used meteorological data obtained from the two runway thresholds. Landing operating minima are usually expressed in terms of a minimum decision height and a minimum runway visual range (RVR) value. RVR is a meteorological parameter measured at the aerodrome, but decision height is not a meteorological variable that can be estimated, as it is a reference for the pilots to decide whether or not continue with the landing. The closer meteorological variable is cloud height (CH), which can be generally found in aerodrome METAR reports. Consequently, we consider the prediction of RVR and CH at the airport. These two variables are critical to determine the acceptable minima for landing operations under different categories of Instrument Landing System (so-called CAT I, CAT II and CAT III). They are also crucial to help airport managers activate low visibility procedures. To obtain the values of these variables, we use direct measurements from three visibilimeters deployed along the airport runway (touchdown zone, the mid-point and stop-end of the runway). These instruments are part of the Meteorological State Agency of the Spanish aeronautical observation network. The complete list of input variables considered in this study are the same for both datasets: hour, temperature in Celsius, relative humidity (\%), wind speed (in KT) and direction (in sexagesimal degrees true) in both runway thresholds (23 and 5 respectively), and atmospheric pressure in hPa. We consider data at the Valladolid airport from winter months (November, December, January and February) of three periods (2009-2010, 2010-2011 and 2011-2012). Hourly values of all the variables are subsequently analysed in this study.

The discretisation of both variables (RVR and CH) follows the simple scheme of Eq. \ref{eq:discretise}, where the thresholds used for RVR are $R_1=300$m, $R_2=550$m and $R_3=2000$m, resulting in four categories. On the other hand, for CH, the discretisation thresholds are $R_1=200$m and $R_2=1500$m, which results in three categories. Note that visibilimeters only deliver precise RVR values when this parameter falls under 2000m. Otherwise the system codifies RVR values as 2000m. This further motivated the use of ordinal classifiers, as the corresponding regression problem would be ill-posed.

According with this discretisation, the number of patterns for the two datasets is included in Table \ref{tab:datasets} along with the class distribution.
RVR data are measured with visibilimeters located along the runway, while, for the estimation of CH, human intervention is needed. This means that RVR is fully available every hour, but the CH information is only available when the airport is open (Valladolid airport is not a 24h airport). That is the reason why the number of available data is different for every variable.

\subsection{Methods tested}

The experimental validation of the methodologies presented in this paper includes the following methods:
\begin{itemize}
	\item Persistence model (Persist), i.e. predicting the label observed in $t$ for time $t+k$. 
	\item A probabilistic autoregressive ordinal model (POM) considering different time windows, which include the previously discussed variables (see Section \ref{sec:data}) from time $t-\Delta$ to time $t$.
	\item The NNPOM method described in Section \ref{sec:nnpom} with the same autoregressive structure. All the methods proposed with Mixture of Experts make use of this classification algorithm for the neural network component. 
	\item Independently Trained Mixture of Experts (ITME), as detailed in Section \ref{sec:indepModel}.
	\item Simultaneously Trained Mixture of Experts (STME), as described in Section \ref{sec:simulModel}, without including specific costs for giving more weight to less frequent classes (i.e. $o_q=1, \forall q$).
	\item The same STME model but including imbalanced costs (STMEIC).
\end{itemize}

\subsection{Experimental setup}

The time series evaluated includes data from $3$ consecutive winters. In order to better validate the methodologies and avoid the dependence of the results on the specific training/test split, we have performed $3$ different splits. For each split, the data from one winter forms the test set, while the other two winters are used for training. Average and standard deviation results are provided.

Different problems were derived according to the prediction time horizon (parameter $k$ in Eq. \ref{eq1}, where we set $k=1$, $k=3$ and $k=6$). Moreover, different input windows were compared, depending on the number of steps before included in the independent variables ($\Delta=1$, $\Delta=3$ and $\Delta=5$). Consequently, a total of $18$ different datasets were included in our experiments ($9$ datasets for each variable). Note that depending on the specific setup some data is not available (for example, given that the CH data is available from 5am, a $\Delta=5$ window means that the first prediction can be done at 10am).

All the models trained by gradient descent methods are stochastic, because the results depend on the initialisation of the parameter vector. Because of this, NNPOM, ITME, STME and STMEIC were run $10$ times, and the results reported are the average performance values of the $10$ final models.

The architecture and training parameters of the neural network models (number of hidden nodes $M$, regularisation parameter $\lambda$ and maximum number of iterations $iter$) have a decisive impact on the performance of the model. Optimal values can vary for each dataset and even for different training/test splits. The most reliable way of fitting these parameters without favouring any method is applying a nested cross-validation procedure and repeating the training process using the value resulting in the best validation performance. In this way, for NNPOM, ITME, STME and STMEIC, a $5$-fold cross-validation model selection was applied, where the ranges explored were: $M\in\{5,10,25,50,75\}$, $iter\in\{100,250,500,1000\}$, $\lambda\in\{0,0.001\}$ (preliminary experiments concluded that, for all datasets, higher regularisation rates always led to worse results).


\subsection{Performance evaluation}

The following performance metrics have been considered in the comparison of models:
\begin{itemize}
	\item The accuracy ($Acc$) is defined by:	
	$$
	Acc= \frac{100}{N} \sum_{i=1}^N [[\hat{y}_i=y_{i}]],
	$$
	where $y_{i}$ is the desired output for time instant $i$, $\hat{y}_i$ is the prediction of the model and $N$ is the total number of patterns in the dataset.
	
	\item The Mean Absolute Error ($MAE$) is a common metric for ordinal classification problems which represents the average deviation in absolute value of the predicted class from the true class (considering the order of the classes in the scale). According to \cite{CruzRamirez2014}, this measure should modified in imbalanced datasets, by taking the relative frequency of the classes into account. In this way, we have evaluated the Average $MAE$ ($AMAE$) and Maximum $MAE$ ($MMAE$) :
	\begin{align}
	AMAE  & = \frac{1}{Q}\sum^{Q}_{q=1}{MAE_{q}} = \frac{1}{Q}\sum^{Q}_{q=1}{\frac{1}{N_q}\sum^{N_q}_{i=1}{e_i}},\nonumber\\
	MMAE & = \frac{1}{Q}\max^{Q}_{q=1}{MAE_{q}},\nonumber
	\end{align}
	where $e_i=|\mathcal{O}(y_i)-\mathcal{O}(\hat{y}_i)|$ is the distance between the true and the predicted ranks, and $\mathcal{O}(\mathcal{C}_q)=q$ is the position of the $q$-th label. $AMAE$ values range from $0$ to $Q-1$, and so do $MMAE$ values.
	
	\item Finally, the geometric mean of the sensitivities of each class ($GMS$) is a summary of the percentages of correct classification individually obtained for each class:
	$$
	GMS = \sqrt[Q]{\prod_{q=1}^{Q}S_q},
	$$
	where the sensitivities, $S_q$, are obtained as:
	\begin{equation*}
	S_{q} = \frac{100}{N_q} \sum_{i=1}^{N_q} [[\hat{y}_i=y_{i}]], q\in\{1,\ldots,Q\},
	\end{equation*}
	where $N_q$ represents the number of patterns of class $\mathcal{C}_q$. This metric is also a standard for imbalanced problems.
\end{itemize}

\begin{table}[htp]
	\caption{Ranking results according to the predictive variable considered (both, CH or RVR). The results for all prediction horizons are averaged.}
	\centering
	\begin{tabular}{rcccc}
		\toprule
		\multicolumn{5}{c}{CH and RVR}\\
		\midrule
		Method & $Acc$ & $AMAE$ & $MMAE$ & $GM$\\
		\midrule
		Persist & $4.22$ & $\mathit{2.28}$ & $\mathit{2.14}$ & $\mathbf{2.17}$\\
		POM & $\mathit{2.89}$ & $4.94$ & $4.94$ & $5.06$\\
		NNPOM & $\mathit{2.89}$ & $4.06$ & $4.06$ & $3.94$\\
		ITME & $3.56$ & $4.50$ & $4.47$ & $4.39$\\
		STME & $\mathbf{1.72}$ & $3.44$ & $3.44$ & $3.17$\\
		STMEIC & $5.72$ & $\mathbf{1.78}$ & $\mathbf{1.94}$ & $\mathit{2.28}$\\
		\midrule
		\multicolumn{5}{c}{CH}\\
		\midrule
		Method & $Acc$ & $AMAE$ & $MMAE$ & $GM$\\
		\midrule
		Persist & $3.78$ & $\mathbf{1.67}$ & $\mathbf{1.39}$ & $\mathbf{1.33}$\\
		POM & $3.89$ & $5.22$ & $5.22$ & $5.44$\\
		NNPOM & $\mathit{3.00}$ & $4.67$ & $4.67$ & $4.56$\\
		ITME & $\mathit{3.00}$ & $3.78$ & $3.83$ & $3.78$\\
		STME & $\mathbf{1.89}$ & $3.78$ & $3.89$ & $3.67$\\
		STMEIC & $5.44$ & $\mathit{1.89}$ & $\mathit{2.00}$ & $\mathit{2.22}$\\
		\midrule
		\multicolumn{5}{c}{RVR}\\
		\midrule
		Method & $Acc$ & $AMAE$ & $MMAE$ & $GM$\\
		\midrule
		Persist & $4.67$ & $\mathit{2.89}$ & $\mathit{2.89}$ & $3.00$\\
		POM & $\mathit{1.89}$ & $4.67$ & $4.67$ & $4.67$\\
		NNPOM & $2.78$ & $3.44$ & $3.44$ & $3.33$\\
		ITME & $4.11$ & $5.22$ & $5.11$ & $5.00$\\
		STME & $\mathbf{1.56}$ & $3.11$ & $3.00$ & $\mathit{2.67}$\\
		STMEIC & $6.00$ & $\mathbf{1.67}$ & $\mathbf{1.89}$ & $\mathbf{2.33}$\\
		\bottomrule
	\end{tabular}
	\label{tab:resultsRankingVariable}
\end{table}

\subsection{Results and discussion}

The mean test results obtained by all the methods compared in this paper can be found in Table \ref{tab:resultsGlobal}. Moreover, 
in order to better summarise these results, Tables \ref{tab:resultsRankingVariable} and \ref{tab:resultsRankingPredictionInstant}
show the test mean rankings in terms of all metrics for all the methods considered in
the experiments. For each
dataset, a ranking of 1 is given to the best method and a 6 is given to the worst one. More specifically, Table \ref{tab:resultsRankingVariable} shows the ranking for the different problems, CH and RVR, and Table \ref{tab:resultsRankingPredictionInstant} shows the results for the considered prediction horizons. 

Several conclusions can be drawn from these results: 
\begin{itemize}
 \item Table \ref{tab:resultsGlobal} shows how stagnant these variables are for the two considered datasets (e.g. RVR being steady 90\% of the time for k=1). Moreover, it can be seen how the prediction of these variables using the persistence model deteriorates for larger time horizons. 
 \item Both Tables \ref{tab:resultsRankingVariable} and \ref{tab:resultsRankingPredictionInstant} show the difficulty in getting a trade-off between all considered metrics, specifically $Acc$ and the rest of metrics. 
 \item The obtained results in Table \ref{tab:resultsGlobal} can be said to be generally satisfactory. For both datasets, we obtain relatively low error in ordinal metrics, such as $AMAE$ and $MMAE$, and good performance both in $Acc$ and $GM$. The performance is usually better for CH than for RVR, which may be because the prediction problem is simpler as there is one class less and the dataset is less imbalanced. Even although both datasets are imbalanced, in most cases no class is completely misclassified ($GM=0.00$). 
 \item Note that the performance gain for larger window sizes is not very high (comparing different $\Delta$ values). This is crucial for CH, given that, the unavailability of results until airport is open limits the first hour when predictions can be obtained, specially for large window sizes.
 \item Comparing POM and NNPOM to Persist in Table \ref{tab:resultsRankingVariable}, it can be seen that in some cases ML models struggle to reach the performance of the persistence. However, NNPOM generally outperforms POM, as expected.
 \item The mixture of models using an independent optimisation of the free parameters (ITME) does not achieve satisfactory results, since in most cases it deteriorates the performance of the persistence model. This can be due to the imbalanced nature of the binary problem solved by the logistic regression (predicting problematic cases), which biases the final model towards the persistence.
 \item The mixture of models that use a simultaneous training shows, however, outstanding performance, being competitive against the persistence model, specially for larger time horizons. The model that includes the imbalanced costs generally presents the best results for $AMAE$, $MMAE$ and $GM$, while the model without costs is competitive in $Acc$. In general, the differences favouring them are higher for the RVR variable, as it is a more difficult problem and there is more room for improvement (see Table \ref{tab:resultsRankingVariable}). Larger prediction horizons ($k=3$ and $k=6$) are also the best scenarios for our proposals (see Table \ref{tab:resultsRankingPredictionInstant}), as the persistence obtains worse results in these cases.
\end{itemize}

\begin{figure}[h!]
	\centering
	\includegraphics[keepaspectratio,width=.5\textwidth]{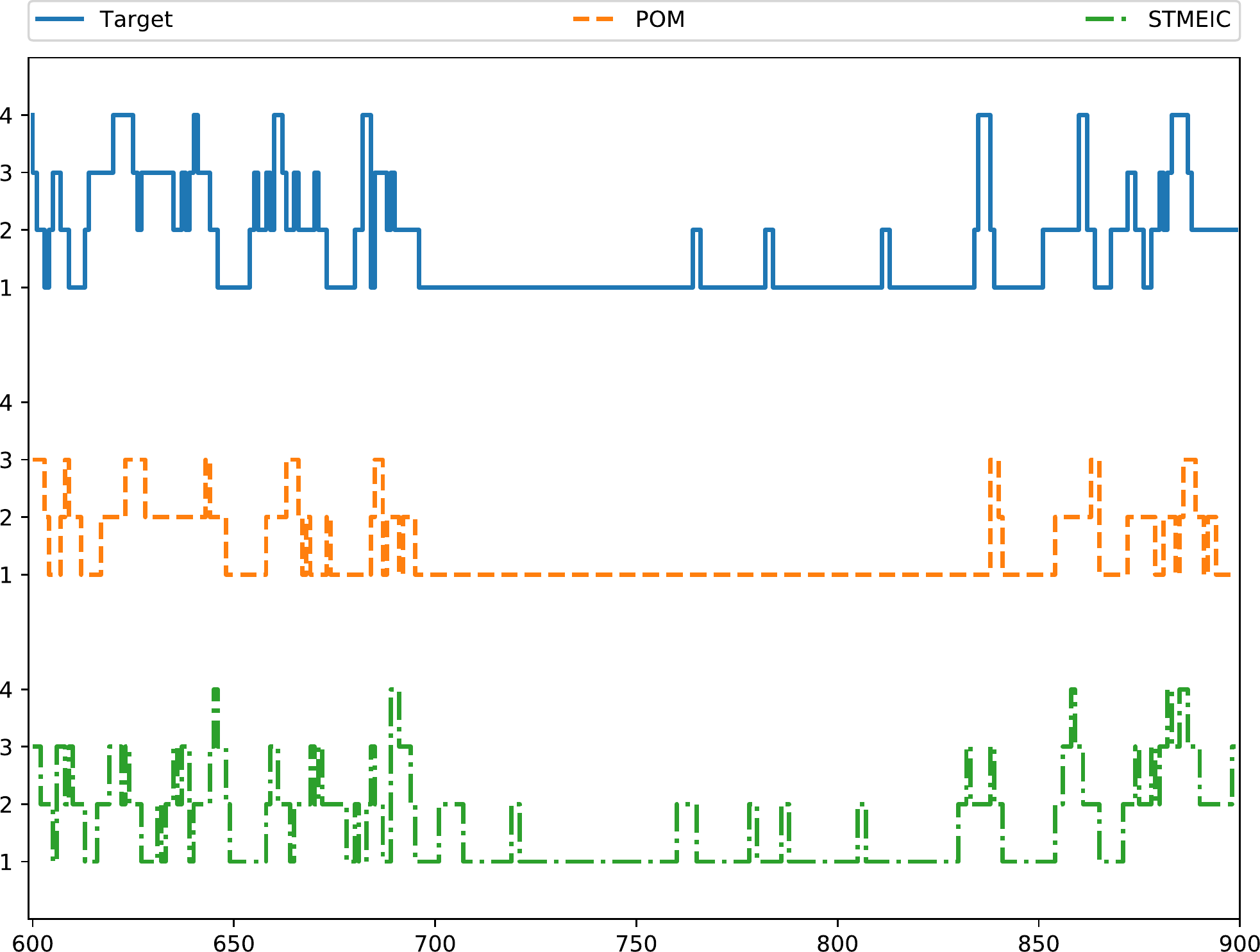}
	\caption{Test target labels for a range of the RVR time series (time horizon $k=3$) and labels predicted by STMEIC and POM (window width $\Delta=3$).}
	\label{fig:predictions}
\end{figure}

The RVR labels obtained by the STMEIC are presented in \figurename{ \ref{fig:predictions}}, compared with target ones and those obtained by POM. As can be seen, the predictions follow the general tendency of the real values, resulting in an acceptably accurate notion of the visibility. STMEIC presents a better prediction than POM for extreme values ($\mathcal{C}_4$) and small fluctuations of visibility.

\begin{table}[htp]
	\caption{Ranking results according to the prediction horizon ($k$). Both variables (CH and RVR) are averaged.}
	\centering
	\begin{tabular}{rcccc}
		\toprule
		\multicolumn{5}{c}{$k=1$}\\
		\midrule
		Method & $Acc$ & $AMAE$ & $MMAE$ & $GM$\\
		\midrule
		Persist & $3.17$ & $\mathbf{1.33}$ & $\mathbf{1.58}$ & $\mathbf{1.33}$\\
		POM & $\mathit{2.67}$ & $3.17$ & $3.17$ & $3.33$\\
		NNPOM & $3.83$ & $5.33$ & $5.33$ & $5.67$\\
		ITME & $3.33$ & $4.00$ & $4.08$ & $3.83$\\
		STME & $\mathbf{2.00}$ & $4.50$ & $4.50$ & $4.17$\\
		STMEIC & $6.00$ & $\mathit{2.67}$ & $\mathit{2.33}$ & $\mathit{2.67}$\\
		\midrule
		\multicolumn{5}{c}{$k=3$}\\
		\midrule
		Method & $Acc$ & $AMAE$ & $MMAE$ & $GM$\\
		\midrule
		Persist & $4.00$ & $\mathit{2.33}$ & $\mathbf{2.00}$ & $\mathbf{2.17}$\\
		POM & $3.17$ & $5.67$ & $5.67$ & $5.83$\\
		NNPOM & $\mathit{2.67}$ & $3.67$ & $3.50$ & $3.00$\\
		ITME & $4.00$ & $4.67$ & $4.67$ & $4.67$\\
		STME & $\mathbf{1.50}$ & $3.17$ & $3.17$ & $3.17$\\
		STMEIC & $5.67$ & $\mathbf{1.50}$ & $\mathbf{2.00}$ & $\mathbf{2.17}$\\
		\midrule
		\multicolumn{5}{c}{$k=6$}\\
		\midrule
		Method & $Acc$ & $AMAE$ & $MMAE$ & $GM$\\
		\midrule
		Persist & $5.50$ & $3.17$ & $2.83$ & $3.00$\\
		POM & $2.83$ & $6.00$ & $6.00$ & $6.00$\\
		NNPOM & $\mathit{2.17}$ & $3.17$ & $3.33$ & $3.17$\\
		ITME & $3.33$ & $4.83$ & $4.67$ & $4.67$\\
		STME & $\mathbf{1.67}$ & $\mathit{2.67}$ & $\mathit{2.67}$ & $\mathit{2.17}$\\
		STMEIC & $5.50$ & $\mathbf{1.17}$ & $\mathbf{1.50}$ & $\mathbf{2.00}$\\
		\bottomrule
	\end{tabular}
	\label{tab:resultsRankingPredictionInstant}
\end{table}

\section{Conclusions}\label{sec:conclusions}

This paper presents a mixture of experts model for predicting ordinal categories associated to low-visibility atmospheric events. Given that these patterns are often persistent in time, the model combines an expert predicting the previous category with an autoregressive neural network expert correcting the persistence when changes are detected. The model is designed from a probabilistic perspective, where the neural network component is based on the proportional odds structure. A gating function, implemented through an autoregressive logistic regression model, assigns the importance of each component.

The model is tested for the task of predicting low-visibility in airports, where the visibility level is represented by two different ordinal categorical variables (cloud height and runway visual height). A battery of experiments is considered, where each variable is evaluated with three different time horizons and three different window widths. Our results are promising, showing very good performance for larger time horizons.

As future research lines, we plan to use more complex neural network models with a recurrent structure to better uncover the dynamics of the time series. Moreover, the training algorithm could be redesigned to alternatively optimise the logistic regression component and the neural network component, in order to accelerate the convergence.


\section*{Acknowledgement}

This work has been subsidized by the projects TIN2014-54583-C2-1-R, TIN2014-54583-C2-2-R, TIN2017-85887-C2-1-P, TIN2017-85887-C2-2-P and TIN2015-70308-REDT of the Spanish Ministry of Economy and Competitiveness (MINECO), FEDER funds (FEDER EU) and the P11-TIC-7508 project of the Junta de Andalucía (Spain).



\bibliographystyle{IEEEtran}
%
\bibliography{references}

\end{document}